%% file: main.tex
\documentclass[journal]{IEEEtran}
\usepackage[T1]{fontenc}
\usepackage{graphicx}
\usepackage{booktabs}
\usepackage{placeins}
\usepackage[round,authoryear]{natbib}
\usepackage[hidelinks]{hyperref}
\graphicspath{{./}}
\begin{document}
\title{Autonomy and Agency in Agentic AI:\\
Architectural Tactics for Regulated Contexts}
\author{
  Damir~Safin and Dian~Balta
  \thanks{D. Safin and D. Balta are with fortiss GmbH, Research Institute of the Free State
  of Bavaria for software-intensive systems, Munich, Germany.
  E-mail: \{safin, balta\}@fortiss.org}
}
\maketitle
\begin{abstract}
\input{abstract}
\end{abstract}
\input{introduction}

\input{background}
\input{design_space}

\input{tactics}

\input{application}
\input{beyond}

\input{conclusion}
\section*{Acknowledgements}
This work was partially supported by the ROBIN project (Grant no: KON-23-039) at the Bavarian Research Institute for Digital Transformation.
\bibliographystyle{IEEEtranN}
\bibliography{main}
\end{document}

%% file: abstract.tex
Deploying agentic AI in regulated contexts requires principled reasoning about two design dimensions: \emph{agency} (what the system can do) and \emph{autonomy} (how much it acts without human involvement). Though often treated independently, they are coupled: at higher autonomy, human error correction is less available, so reliable operation requires constraining agency accordingly; compliance requirements reinforce this by mandating human involvement as action consequences grow. Yet no established approach addresses them jointly, leaving practitioners without a principled basis for reasoning about oversight, action consequences, and error correction.

This work introduces a two-dimensional design space in which both dimensions are organised into five operational levels, making the coupling explicit and navigable. Autonomy ranges from human-commanded operation (L1) to fully autonomous monitoring (L5); agency ranges from reasoning over supplied context (L1) to committed writes to authoritative records (L5). Building on this space, we propose six architectural tactics---checkpoints, escalation, multi-agent delegation, tool provisioning, tool fencing, and write staging---for adjusting a deployment's position within it. The tactics are grounded in two worked examples from public-sector contexts, illustrating how they apply under realistic compliance constraints. We further examine five deployment parameters---model capability, agent architecture, tool fidelity, workflow bottlenecks, and evaluation---that shape what is achievable at any configuration independently of agency and autonomy. Together, the design space, tactics, and deployment parameters provide a shared vocabulary for principled, compliance-aware agentic AI design in which responsibility, auditability, and reversibility are explicit design considerations rather than properties that must be retrofitted after deployment.

%% file: introduction.tex
\section{Introduction}\label{sec:introduction}

Advances in large language models (LLMs) have catalysed a shift from static AI assistants toward \emph{agentic} systems~\citep{Baird2021} capable of multi-step planning, tool use, and autonomous action across complex workflows~\citep{AbouAli2025}. In regulated deployment contexts, this shift introduces a distinctive accountability challenge: as agents act with greater independence, the chain of responsibility between a decision and a human decision-maker grows longer and more opaque. Consequential actions become harder to attribute, audit trails harder to maintain, and errors more difficult to reverse before they propagate. Data protection regulations such as the GDPR~\citep{gdpr_2016} and the EU AI Act~\citep{eu_ai_act_2024} reinforce this by mandating traceability and human oversight over automated decisions.

\citet{Feng2025Levels} treat autonomy and agency as independent design dimensions, decomposing autonomy into five escalating levels and distinguishing it from agency---the scope of what an agent can perceive and affect. Yet in regulated deployment, the two cannot be navigated independently: a highly capable agent without appropriate oversight may breach accountability norms; one that is too tightly constrained offers negligible practical benefit~\citep{Grnsund2020}. Missing is a systematic account of the architectural tactics available for balancing both dimensions in practice. The gap is consequential because the dimensions are coupled: at higher autonomy, human error correction is less available, so reliable task completion requires limiting agency to what the model can handle independently. Without such tactics, practitioners can describe where a deployment sits in the design space but lack principled means to navigate it.

This work makes three contributions:
\begin{enumerate}
    \item \textbf{Design space.} A two-dimensional space in which Autonomy (L1--L5) and Agency (L1--L5) are each organised into five operational levels, making the coupling between the dimensions explicit and navigable.
    \item \textbf{Architectural tactics.} Six tactics---three targeting autonomy (Checkpoints, Escalation, Multi-Agent Delegation) and three targeting agency (Tool Provisioning, Tool Fencing, Write Staging)---for adjusting a deployment's position within the space, grounded in two worked examples.
    \item \textbf{Deployment parameters.} Five parameters---model capability, agent architecture, tool fidelity, workflow bottlenecks, and evaluation---that shape the achievable ceiling at any design-space position, independently of the agency--autonomy configuration.
\end{enumerate}
Together, these contributions provide a shared vocabulary for compliance-aware agentic AI design in which responsibility, auditability, and reversibility are explicit design considerations.

This work is organised as follows. Section~\ref{sec:background} situates the work within the literature on agentic AI, autonomy and agency, and regulatory constraints. Section~\ref{sec:design_space} introduces the two-dimensional design space and four representative configurations. Section~\ref{sec:tactics} proposes the six architectural tactics and discusses how they can be selected and combined. Section~\ref{sec:application} applies the tactics in two worked examples. Section~\ref{sec:beyond} examines the five deployment parameters that shape the achievable ceiling of the design space. Section~\ref{sec:conclusion} concludes and identifies directions for future work.

%% file: background.tex
\section{Background}\label{sec:background}

\subsection{Agentic AI Systems}

Agentic AI systems are characterised by their ability to perform multi-step planning, invoke external tools, maintain persistent memory, and coordinate action across dynamic environments~\citep{AbouAli2025}; a defining feature is their capacity to decompose complex goals and orchestrate networks of specialised sub-agents to accomplish objectives that exceed individual model capacity~\citep{Sapkota2026}. In the context of intelligent automation, LLMs have shown particular promise in handling unstructured inputs and generating adaptive workflows, addressing limitations of earlier rule-based approaches~\citep{Sonnabend2025}. Despite these capabilities, agentic systems remain imperfect---susceptible to hallucination, adversarial manipulation, and misaligned action~\citep{Winston2025}---prompting research into human-in-the-loop architectures that combine oversight with autonomous efficiency~\citep{Mozannar2025}. A systems-level perspective is essential: emergent behaviour arising from agent interaction with environments and other agents cannot be captured by evaluating individual model capabilities alone~\citep{Miehling2025}. While this body of work characterises what agentic systems can do and where they fall short, it does not provide systematic design guidance for positioning them along capability and oversight dimensions in regulated deployment contexts.

\subsection{Autonomy and Agency}

\citet{Feng2025Levels} treat autonomy and agency as two distinct and independent design dimensions. Autonomy is defined as the extent to which an agent is \emph{designed} to operate without user involvement---a deliberate design decision, not a capability property---and is decomposed into five escalating levels characterised by the role a user takes: from \emph{Operator} (human commands, agent executes a specific sub-task) through \emph{Collaborator}, \emph{Consultant}, and \emph{Approver}, to \emph{Observer} (agent operates fully autonomously under human monitoring). Agency is defined as the capacity to formulate an intention for an action and carry it out; operationally, it is shaped by the set of tools available to the agent. Critically, the two dimensions vary independently: an agent with limited tools but no user supervision has low agency and high autonomy, while an agent with broad tool access that frequently solicits user feedback has high agency but low autonomy.

How human involvement should be configured has been examined from design, empirical, and risk perspectives. Interaction design patterns such as editable agent memory and explicit approval steps embed oversight requirements directly into the system surface~\citep{Feng2025Interfaces}. Empirical work underscores the stakes of these choices: fully delegated AI decision-making is perceived as procedurally less fair than ensemble modes in which a human remains jointly involved, and this fairness deficit erodes trust in the responsible official~\citep{Diebel2025}. More broadly, \citet{Mitchell2025} argue that risks to people increase monotonically with autonomy and that semi-autonomous configurations---which preserve meaningful human oversight---offer a more favourable risk-benefit profile than fully autonomous agents. Research on generative AI in knowledge work identifies further tensions---between productivity and reflection, and between availability and reliability---for which human-in-the-loop designs are proposed as partial remedies~\citep{Gie2025}. Research on trust and reliance confirms that appropriate configurations are context-dependent: human-related, AI-related, and decision-related factors all modulate when reliance is appropriate, and no single configuration generalises across deployment contexts~\citep{Spatscheck2025}. While \citet{Feng2025Levels} establish autonomy and agency as independent design dimensions and propose five autonomy levels, neither that work nor the surrounding literature provides a systematic account of the architectural tactics available for adjusting position along either axis.

\subsection{Regulatory and Compliance Constraints}

Deploying agentic AI in regulated contexts introduces constraints that governance research has begun to map. The EU AI Act~\citep{eu_ai_act_2024} classifies certain AI applications as high-risk and mandates specific human oversight requirements; the GDPR~\citep{gdpr_2016} constrains data access and grants individuals rights to explanation over automated decisions affecting them. For agentic AI in public-sector organisations specifically, existing oversight mechanisms---designed around episodic approvals and siloed compliance units---are challenged by the continuous, cross-departmental nature of agent operation; addressing this requires integrating governance into operational design rather than treating it as a post-hoc compliance layer~\citep{Schmitz2025}. More broadly, accountability in human-AI agent relationships requires conditional engagement mechanisms rather than reliance on retrospective audit~\citep{Lange2025}. The governance literature characterises which constraints apply in regulated contexts but does not specify how they bound the agency--autonomy design space---which configuration regions remain viable and which exceed compliance limits.

%% file: design_space.tex
\section{Autonomy and Agency as Design Dimensions}\label{sec:design_space}

Our analysis is structured around a two-dimensional design space in which \emph{autonomy} and \emph{agency} form orthogonal axes, each decomposed into five operational levels. We use this space to identify example configurations and to reason about configuration viability given compliance constraints. The five autonomy levels are adapted from \citet{Feng2025Levels}; the five agency levels are introduced in this work, following the same decomposition approach.

\subsection{Core Dimensions}

Two design dimensions jointly characterise the role and reach of an agentic system in complex workflows, separating two qualitatively distinct classes of design decision.

\noindent\textit{\textbf{Autonomy.}}
Autonomy describes \emph{how much human involvement is required}: it is a decision about the user's role in the loop~\citep{Feng2025Levels}. The spectrum ranges from systems in which the human drives every action (low autonomy) to systems that operate fully independently under monitoring (high autonomy). Lower autonomy is not a deficiency; in high-stakes workflows it is often an intentional design choice that enforces checkpoints, review steps, and human decision authority, thereby satisfying accountability and auditability requirements.

\noindent\textit{\textbf{Agency.}}
Agency is operationalised here as \emph{what the system can perceive and affect}: defined jointly by the \emph{environment} (tools and effectors available to the agent) and the \emph{context} (information and data it can access). In practice, what an agent can do is constrained as much by information access as by action scope. Agency is also task-relative: adding a tool or data source increases agency only if it expands what the system can accomplish for the specific task at hand; redundant additions that do not change the reachable outcome space do not meaningfully increase agency.

In both dimensions, compliance requirements---including the GDPR~\citep{gdpr_2016}, the EU AI Act~\citep{eu_ai_act_2024}, and internal security policies---set hard limits: on agency, by restricting which data and systems the agent may access; on autonomy, by reserving certain decisions for human decision-makers.

\subsection{Levels of Autonomy}

The five levels of autonomy, adapted from \citet{Feng2025Levels}, describe the \emph{interaction design}: how control, review, and responsibility are allocated between the human and the agent. They function as oversight and safety controls, determining where checkpoints are placed in a workflow, who is accountable for decisions, and how errors are surfaced and corrected. Table~\ref{tab:autonomy} summarises the five levels.

\begin{table*}[ht]
    \centering
    \caption{Levels of Autonomy for agentic AI, adapted from \citet{Feng2025Levels}.}\label{tab:autonomy}
    \small
    \setlength{\tabcolsep}{4pt}
    \begin{tabular}{lp{2cm}p{5cm}p{5.5cm}}
        \toprule
        \textbf{Level} & \textbf{Stance} & \textbf{Interaction Design} & \textbf{Example} \\
        \midrule
        L1 & Operator     & Human commands; agent executes a specific sub-task. & Official queries relevant policy; agent retrieves and summarises the applicable rules. \\[4pt]
        L2 & Collaborator & Shared workspace; human and agent iterate back-and-forth. & Agent drafts a permit assessment; official edits sections and assigns sub-checks. \\[4pt]
        L3 & Consultant   & Agent leads planning; pauses to confirm direction with the human. & Agent outlines a regulatory review plan and asks the official to confirm which statutes to prioritise. \\[4pt]
        L4 & Approver     & Agent executes; pauses for explicit approval before consequential actions. & Agent prepares a complete case file and awaits sign-off before writing to the record. \\[4pt]
        L5 & Observer     & Agent operates fully autonomously; human monitors and can intervene. & Automated compliance checks run end-to-end; supervisor reviews flagged exceptions. \\
        \bottomrule
    \end{tabular}
\end{table*}

\subsection{Levels of Agency}

The five levels of agency provide a practical scale of increasing operational reach. The levels are cumulative: a higher level typically incorporates the capabilities of all lower levels. Table~\ref{tab:agency} presents the five levels. A critical distinction is that reversibility is an \emph{agency} property of available actions---determined by tool semantics and practical rollback cost---not a function of who authorises the action. Whether a human must approve an action is determined by the autonomy design; whether the action can be undone is a property of the action itself.

\begin{table*}[ht]
    \centering
    \caption{Levels of Agency for agentic AI, introduced in this work following the same decomposition approach as \citet{Feng2025Levels}.}\label{tab:agency}
    \small
    \setlength{\tabcolsep}{4pt}
    \begin{tabular}{lp{2.5cm}p{5cm}p{4.5cm}}
        \toprule
        \textbf{Level} & \textbf{Agent Type} & \textbf{Agency Scope} & \textbf{Example} \\
        \midrule
        L1 & Model-only             & Reason, draft, and summarise from provided context; no retrieval or tool use. & Agent rewrites a policy section in plain language from text supplied by an official. \\[4pt]
        L2 & Internal agent         & Safe local tools (validators, sandboxed code) with no access to external systems. & Sandboxed ruleset flags missing fields and date conflicts in a draft form. \\[4pt]
        L3 & Read-only agent        & Read-only access to external sources (web, repositories, read-only APIs); cannot write or modify records. & Agent searches a case-law database and returns a structured memo with citations. \\[4pt]
        L4 & Pragmatic (reversible) & External writes with low rollback cost: drafts, pending records, editable fields. & Agent creates a pending case note that a caseworker can review and amend before submission. \\[4pt]
        L5 & Pragmatic (commit)     & External writes with costly undo: final submissions, authoritative record updates, payments. & Agent submits a completed filing; reversal requires formal cancellation or appeal. \\
        \bottomrule
    \end{tabular}
\end{table*}

\subsection{Navigating the Design Space}

\begin{figure*}[ht]
    \centering
    \includegraphics[width=0.8\textwidth]{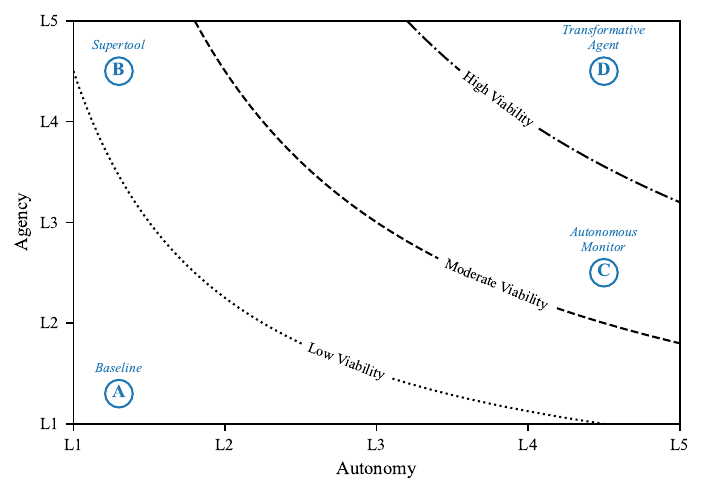}
    \caption{Sketch of operational viability across Autonomy and Agency dimensions. Iso-viability curves indicate configurations of equal viability; no formal relationship is assumed---they convey that one configuration is qualitatively preferable to another for a given deployment context. Points A--D mark four example configurations.}
    \label{fig:design_space}
\end{figure*}

A central design tension lies between an agent's capacity to act and the degree of human oversight required. The two-dimensional design space provides an analytical frame for reasoning about this tension: a given configuration occupies a position in the space, and the question of whether that position is viable in a specific deployment context is captured by its \emph{operational viability}---a qualitative indicator of whether one configuration is preferable to another under the applicable compliance constraints.

As illustrated in Figure~\ref{fig:design_space}, configurations of equal operational viability form \emph{iso-viability curves}: any design that falls along the same curve is equally preferable from a viability standpoint. The curves reflect a practical coupling driven by task reliability: an agent operating at higher autonomy has less opportunity for human error correction~\citep{Feng2025Levels,Mitchell2025,anthropic2026agents}, so reliable task completion is feasible only when agency is constrained to what the model can handle independently. The relationship is bidirectional---increasing autonomy compresses the viable agency range; accepting lower autonomy opens it. Compliance requirements reinforce this coupling independently: the EU AI Act mandates human oversight for high-risk applications; the GDPR constrains data access and grants rights to explanation over automated decisions~\citep{eu_ai_act_2024,gdpr_2016}.

Four example configurations illustrate positions in the design space:

\noindent\textit{\textbf{(A) Baseline.}}
At low agency and low autonomy, this region represents fully manual, human-driven task execution. Because the operational friction is high and leverage is low, systems in this region have negligible operational viability as an agentic deployment. \textit{Illustration:} a human who supplies document text and manually transfers the output to a downstream system; the agent responds with a plain-language summary but performs no retrieval, no writes, and no tool use.

\noindent\textit{\textbf{(B) Supertool.}}
At high agency and low autonomy, systems in this region possess powerful, often irreversible capabilities. The system is designed with low autonomy so the human acts as the orchestrator, resulting in a highly viable human-orchestrated actuator that maximises impact while satisfying stringent compliance constraints. \textit{Illustration:} a human operator who triggers each step explicitly; the agent executes the operation---updating a database record or dispatching a formal notification---on request.

\noindent\textit{\textbf{(C) Autonomous Monitor.}}
At read-only agency and high autonomy, systems operate with full or near-full autonomy but are strictly confined to read-only access to external sources. Because the agent cannot alter any system of record, the risk of consequential wrong action is minimised, making this a viable high-autonomy configuration even under stringent compliance constraints. \textit{Illustration:} an agent that continuously scans new legislation, generates structured briefings, and posts them to an internal dashboard; the human supervisor reads the output and monitors for exceptions.

\noindent\textit{\textbf{(D) Transformative Agent.}}
At high agency and high autonomy, this is the theoretical high-value target of agentic deployment, where a system can autonomously execute highly consequential actions. While this region offers maximum leverage and efficiency, it currently represents the highest barrier to practical deployment due to technical limitations, hallucination risks~\citep{Huang2024}, and rigid legal and regulatory requirements. \textit{Illustration:} an agent that autonomously executes all steps of a procurement or contract filing process, writing to authoritative records at every step; the human monitors and may intervene but does not sign off on individual actions.

%% file: tactics.tex
\section{Architectural Tactics}\label{sec:tactics}

The design space characterises \emph{where} a system sits; this section addresses the complementary question of \emph{how} it can be moved along either dimension. We propose six architectural tactics, grouped by the axis they primarily affect, and summarised in Table~\ref{tab:tactics}. Because the dimensions are coupled, tactics from both axes can be combined: reducing agency through fencing or write staging lowers the risk profile and may permit higher autonomy without exceeding compliance bounds; when high agency is required by the task, autonomy tactics that introduce human oversight at consequential steps can bring the overall configuration within those bounds.

\begin{table*}[ht]
    \centering
    \caption{Classification of architectural tactics. Implementation layer indicates where the tactic operates (workflow, tool, or action level); direction of effect indicates whether the tactic raises or lowers the target dimension; diagnostic question identifies the key deployment condition that makes the tactic applicable.}\label{tab:tactics}
    \small
    \setlength{\tabcolsep}{4pt}
    \begin{tabular}{lp{2.8cm}lllp{5.8cm}}
        \toprule
        \textbf{ID} & \textbf{Tactic} & \textbf{Axis} & \textbf{Layer} & \textbf{Effect} & \textbf{Diagnostic question} \\
        \midrule
        T1 & Checkpoints            & Autonomy & Workflow & $\downarrow$         & Is review at this step warranted unconditionally, regardless of output quality? \\[2pt]
        T2 & Escalation             & Autonomy & Workflow & $\downarrow$         & Does reliability vary across inputs, and can low-confidence cases be detected at runtime? \\[2pt]
        T3 & Multi-Agent Delegation & Autonomy & Workflow & $\uparrow$           & Can a sub-task be handled reliably by a specialised agent without human sign-off? \\
        \midrule
        T4 & Tool Provisioning      & Agency   & Tool     & $\uparrow\downarrow$ & Is the tool set mismatched to the task---missing required capabilities, or containing tools the task does not need? \\[2pt]
        T5 & Tool Fencing           & Agency   & Tool     & $\downarrow$         & Does a required tool expose more operations or reachable targets than the task warrants? \\[2pt]
        T6 & Write Staging          & Agency   & Action   & $\downarrow$         & Does the workflow write to an authoritative record, and is that write currently irreversible? \\
        \bottomrule
    \end{tabular}
\end{table*}

\subsection{Tactics Targeting Autonomy}

\noindent\textit{\textbf{T1: Checkpoints.}}
A checkpoint is a fixed, workflow-bound pause~\citep{Mozannar2025}: the defining characteristic is that the pause fires unconditionally at a predetermined step, regardless of the agent's output quality or confidence. When control passes to the human, they may approve and continue, reject and abort, or correct the agent's output before the workflow proceeds---flexibility that gives checkpoints their accountability value while the unconditional trigger makes review locations auditable by design. Inserting a checkpoint moves the effective autonomy level toward L4 (Approver) at that step; removing one moves it toward L5 (Observer). Checkpoints are appropriate when the need for oversight at a step is known in advance and applies regardless of the agent's output---that is, when a fixed review point is warranted by design rather than contingent on what the agent produces.

\noindent\textit{\textbf{T2: Escalation.}}
Escalation is a dynamic, agent-initiated pause at the workflow level: the defining characteristic is that the agent decides when to stop, based on its own assessment of the situation. It fires when the agent determines it cannot or should not proceed autonomously---for example, when confidence in a retrieved answer falls below a threshold~\citep{Wang2026}, when inputs are ambiguous or contradictory, or when a situation lies outside the agent's sanctioned scope---lowering the effective autonomy at the affected step and restoring human involvement at the point where the agent reaches its reliable operating boundary. Because escalation is conditional, it allows a system to operate at high autonomy in the common case while preserving a reliable fallback path for edge cases. Escalation is appropriate when the agent's reliability across inputs is uneven and the need for human involvement cannot be specified in advance. Poorly calibrated escalation---too frequent or too vague---degrades user experience and erodes trust in the system.

\noindent\textit{\textbf{T3: Multi-Agent Delegation.}}
Multi-agent delegation~\citep{Sapkota2026} is a workflow-level tactic that increases effective autonomy by replacing a human sub-task handler with a specialised sub-agent. Where escalation transfers a step to a human when the agent reaches a limit, delegation transfers it to another agent that operates autonomously within its own scope. The orchestrating agent decomposes a workflow step into a sub-task, dispatches it to a sub-agent equipped with appropriate tools and context, and incorporates the result without human involvement---raising the effective autonomy level of the overall system. The tactic is applicable where the sub-task is sufficiently well-defined for a specialised agent to handle reliably, and where the compliance context permits automated handling of that sub-task without human sign-off. A critical accountability implication is that errors introduced by a sub-agent are harder to attribute than errors in a direct human decision, making audit trail preservation and scope constraint across the delegation boundary as important as for the orchestrating agent itself.

\subsection{Tactics Targeting Agency}

\noindent\textit{\textbf{T4: Tool Provisioning.}}
The set of tools made available to the agent at runtime is the most direct lever for controlling agency~\citep{alampara2025task}. Provisioning a new tool raises the agent's agency level if the tool expands the reachable outcome space for the task at hand; revoking a tool lowers it. In practice, tool sets are often defined per deployment context rather than per agent type: the same underlying model may be given read-only database access in one deployment (L3) and write access in another (L4--L5), producing different effective agency levels without any change to the model itself. The tactic is applicable whenever the tool set is mismatched to the task---either because required capabilities are absent and must be provisioned, or because tools present in the set exceed what the task warrants and should be revoked.

\noindent\textit{\textbf{T5: Tool Fencing.}}
Tool fencing constrains the operational envelope of a tool without removing it from the agent's available set. It takes two forms. Resource-level fencing restricts which targets a tool can address through its configuration---for example, limiting a retrieval tool to a specific document corpus. Parameter-level fencing restricts the values or combinations of values the tool may accept at call time---for example, a write tool whose payload is validated against a schema that excludes fields reserved for human entry. Both forms are implemented in the tool's own validation layer, making the constraint independent of the model's behaviour: even if the model attempts an out-of-bounds call, the tool rejects it before any external action is taken. This makes fencing a reliable agency control that does not depend on prompt engineering or model-level instruction-following~\citep{Debenedetti2025}. It is applicable where a tool is appropriate for the task but its full operational surface---whether in terms of reachable targets or permitted parameter values---exceeds what the task or compliance context warrants~\citep{Errico2025}.

\noindent\textit{\textbf{T6: Write Staging.}}
Write staging interposes a draft or pending layer between an agent action and its authoritative commit, converting an otherwise irreversible L5 write to a reversible L4 action until the staged output is explicitly promoted. It is an agency tactic because it operates on the reversibility of the action itself, not on the workflow control structure: the same staging layer applies regardless of whether the surrounding workflow uses a checkpoint, an escalation, or neither. The tactic is applicable when the compliance or risk context requires reversibility and the underlying system of record supports a two-phase write---a propose step that creates a provisional record, and a commit step that makes it authoritative~\citep{Mohammadi2026}. Where native draft support is absent, an external staging layer with a human-initiated commit step is an alternative, at the cost of added latency. In practice, write staging arises wherever an agent writes to a system of record that does not support undo: the agent writes to a provisional record rather than directly to the authoritative system; if the action is wrong, the draft is discarded and the authoritative record remains untouched.

\subsection{Selecting and Combining Tactics}

The choice among the autonomy tactics turns on whether the need for human involvement is predetermined, conditional, or eliminable. Checkpoints are appropriate where oversight is applied at a known workflow step regardless of output quality, making the review location fixed and auditable by design. Escalation is appropriate where the need for human involvement is conditional on the agent's own assessment and cannot be specified in advance. Multi-agent delegation is applicable where a sub-task is sufficiently well-defined to be handled by a specialised agent without human sign-off. The three are not mutually exclusive: a single workflow may apply checkpoints at high-stakes steps, escalation at ambiguous edge cases, and delegation for well-scoped sub-tasks.

Among the agency tactics, the distinctions are operational. Tool provisioning and revocation determine what the agent can reach and over what target set; tool fencing restricts how the agent may parameterise a call to an already-available tool. Fencing is a lower-cost mechanism than revocation when the required constraint is at the call level rather than the resource level. Write staging addresses reversibility and is applicable wherever the target system supports or can be extended with a two-phase write; it is orthogonal to provisioning, revocation, and fencing, and may be combined with any of them.

Because autonomy and agency are coupled in the design space, adjustments on one axis affect the viable range on the other. Reducing agency---through revocation, fencing, or write staging---lowers the risk profile of the configuration and may permit higher autonomy without exceeding compliance bounds. Conversely, when high agency is required by the task, autonomy tactics that introduce human oversight at consequential steps can bring the overall configuration within those bounds. Tactics from both axes may be freely combined within a single deployment.

%% file: application.tex
\section{Worked Examples}\label{sec:application}

This section applies the design space and tactics to two concrete deployments in different domains, illustrating how system architects can reason about configuration choices and compliance bounds in their own contexts.

\subsection{Public-Sector Document Classification}\label{sec:application_docclass}

\subsubsection{Deployment Context}

A public-sector organisation processes a continuous stream of incoming items---PDFs, emails with attachments, and receipts. The task is to classify each item into the responsible organisational unit, summarise it, and dispatch it by email. The organisation operates an established document management infrastructure with a web UI and API, through which domain experts manage the classification task. Data protection obligations constrain the agent's data access; accountability requirements mandate that classification decisions remain auditable. The task's repetitive, high-volume character, well-defined success criterion, and the presence of human recipients at the end of the dispatch chain made high autonomy operationally attractive; starting from configuration~A, the design goal was to move toward~D---full end-to-end autonomy with the level of agency the task required---within the bounds the compliance context permitted.

\subsubsection{Tool Provisioning and Tool Fencing (T4, T5)}

The existing infrastructure was reused directly, with \emph{Tool Fencing}~(T5) applied to limit the agent's reachable operations to those the task required: the agent read case records and retrieved document metadata through the API, but write operations outside the classification and dispatch workflow were fenced out. Two capabilities were not available through the API: document-to-text parsing and outbound email dispatch. Both were implemented as dedicated tools and added to the agent's tool set via \emph{Tool Provisioning}~(T4); the email-dispatch tool in particular moved the system from read-only to pragmatic agency, requiring the autonomy configuration to be designed accordingly.

\emph{Write Staging}~(T6) was considered but not applied. The email-dispatch action sat at Agency~L4---pragmatic reversible---rather than L5, and the presence of human recipients at the end of the dispatch chain provided a natural error-correction layer. Had the system written directly to an authoritative case record rather than to a human recipient, T6 would have been warranted.

\subsubsection{Multi-Agent Delegation (T3)}

The agent's orchestration layer was structured around the way domain experts approached the classification task: they read the document title and skimmed the body before reaching a decision. The system reflected this through a dedicated summarisation sub-agent producing a structured summary of each item, and a second agent combining this summary with the output of the existing ML classifiers to produce a natural-language justification for the proposed classification.

Both sub-agents were instances of \emph{Multi-Agent Delegation}~(T3), applied because each sub-task was well-scoped enough to be handled reliably without human involvement. The two sub-agents served distinct purposes: the summarisation sub-agent improved classification fidelity; the justification sub-agent ensured a natural-language explanation was produced consistently for every item, independent of which autonomy configuration was active---preserving the audit trail across both checkpoint and autonomous dispatch modes.

\subsubsection{Checkpoints and Escalation (T1, T2)}

The central design decision for the dispatch step was whether to require human approval before the email-dispatch tool was invoked, or to allow autonomous dispatch with a conditional fallback.

A \emph{Checkpoint}~(T1) at this step required a domain expert to review the classification, summary, and justification through the web UI before confirming dispatch. As classifier accuracy improved, the expected frequency of misdirected items fell below the threshold at which unconditional oversight was warranted, and the checkpoint was omitted from the production configuration.

In its place, a conditional \emph{Escalation}~(T2) mechanism was applied: when the classifier's confidence fell below a defined threshold, the item was routed to a domain expert for review rather than dispatched autonomously.

\bigskip\noindent Mapped onto the design space: \emph{Tool Provisioning}~(T4) placed the configuration at Agency~L4 via the email-dispatch tool; \emph{Tool Fencing}~(T5) kept the API scope narrow, preventing write access beyond classification and dispatch; \emph{Multi-Agent Delegation}~(T3) provided consistent summarisation and justification, supporting reliable autonomous operation. \emph{Escalation}~(T2) enabled Autonomy~L5 in the high-confidence regime, with low-confidence items dropping to Autonomy~L4 when routed to a domain expert for review.

\subsection{Automated Funding Program Matching}\label{sec:application_funding}

\subsubsection{Deployment Context}

An automated funding program matching tool is designed to support citizen-driven innovation. Public funding landscapes are fragmented across thousands of programs whose eligibility criteria are expressed in natural language; applicants---individuals, small organisations, community initiatives---must navigate this complexity without requiring legal expertise. The system separates two concerns: an offline stage that compiles natural-language program criteria into executable eligibility rules, and a query stage that matches a project description submitted by the applicant against those rules at runtime.

\subsubsection{Tool Provisioning and Tool Fencing (T4, T5)}

In the offline stage, an agent writes executable eligibility rules to the program database, placing the system at Agency~L4. \emph{Tool Provisioning}~(T4) equipped the query stage with a geodata API that resolves postcodes and place names into standardised geographic identifiers. \emph{Tool Fencing}~(T5) constrains the generated eligibility rules to a sandboxed execution environment, preventing LLM-generated logic from accessing external systems.

\subsubsection{Checkpoints and Escalation (T1, T2)}

Before the search proceeds, a \emph{Checkpoint}~(T1) presents the applicant with the structured project profile derived from their project description---funding amount, location, project type---for review and correction. \emph{Escalation}~(T2) handles programs for which a clear eligibility verdict cannot be reached: rather than forcing a decision, the system presents these programs to the applicant with an explicit indication that eligibility could not be confirmed, leaving the final judgement to them.

\medskip\noindent Mapped onto the design space: the offline rule-generation stage operates at Agency~L4 with Autonomy~L5---write access to the program database with no per-rule human sign-off. \emph{Tool Provisioning}~(T4) extended the query stage with a geodata API; \emph{Tool Fencing}~(T5) bounds the rule executor to a sandboxed environment, preventing the generated logic from reaching external systems; \emph{Checkpoint}~(T1) moves the query-stage autonomy to L4 at the profile-review step; \emph{Escalation}~(T2) restores L4 autonomy for ambiguous eligibility cases while allowing L5 in the clear-verdict regime.

\subsection{Comparison Across Examples}

The two worked examples illustrate how the same set of tactics applies differently depending on the deployment context. In document classification, the primary design challenge was enabling high autonomy at moderate agency while preserving auditability; the tactics applied---multi-agent delegation, escalation, tool fencing---addressed reliability and accountability in a high-volume, internally-facing workflow. In funding program matching, the challenge was managing the risk of irreversible rule generation while giving the applicant appropriate control; the tactics applied---checkpoints, escalation, tool fencing---prioritised transparency and user oversight in an externally-facing, lower-volume context. In both cases, the design space provides a common vocabulary for characterising the configuration and the compliance constraints that bound it.

%% file: beyond.tex
\section{Beyond Agency and Autonomy}\label{sec:beyond}

The model and tactics presented in the preceding sections treat agency and autonomy as the primary design variables, holding other system parameters fixed. This is a deliberate simplification: isolating the two axes makes the trade-offs tractable and the design space navigable. In practice, however, several additional parameters shape the operational viability of a system independently of where it sits on those axes. These parameters determine the ceiling of what is achievable at any given agency--autonomy position---they shift the iso-viability curves rather than move the system along them. The following discussion is illustrative rather than exhaustive, and the parameters are not presented in order of importance: their relative weight is task- and context-dependent. These parameters are most effectively considered alongside, not after, the agency and autonomy design decisions.

\noindent\textit{\textbf{Model Capability.}}
The reasoning and instruction-following quality of the underlying LLM is a full design decision: the model is chosen, not inherited. A more capable model tolerates higher input ambiguity, commits fewer tool-use errors, and sustains reliable behaviour at higher agency levels. Concretely, a task that requires L4 human oversight under a weaker model may be safely delegated to L5 autonomy under a stronger one, because the error rate falls below the threshold at which human correction becomes necessary. Model capability thus shifts the viable region of the design space: stronger models move the iso-viability curves toward the origin, making configurations accessible that were previously out of reach.

\noindent\textit{\textbf{Agent Architecture.}}
Beyond the choice of model, the internal structure of the agentic system---how it manages memory, whether it performs explicit multi-step planning, and whether it is composed of multiple specialised agents---determines whether the agent can actually perform the task assigned to it at a given agency level. An agent without persistent memory cannot maintain coherent context across a multi-step workflow; one without planning cannot decompose a complex case into actionable sub-tasks~\citep{Sapkota2026}. These are design decisions with significant impact on achievable operational viability, and they are largely orthogonal to the agency and autonomy settings: two systems with identical tool sets and identical autonomy configurations may perform very differently depending on their internal architecture.

\noindent\textit{\textbf{Tool Fidelity.}}
Tool fidelity---the reliability and accuracy of the tools the agent uses---is partly a design decision and partly an inherited constraint. In many deployment contexts, the relevant systems of record, APIs, and databases are often already established; their quality is not freely chosen by the system designer. A retrieval tool that returns stale or imprecise results limits what is achievable at agency L3 regardless of how the agent is configured; an unreliable write API makes L4--L5 deployments risky independent of the autonomy design. Where tool fidelity cannot be improved directly, it functions as a constraint that bounds the viable region of the design space.

\noindent\textit{\textbf{Workflow Bottlenecks.}}
Tasks such as application processing, approval decisions, or eligibility assessment are not atomic actions but chains of steps: intake, verification, analysis, decision, and notification, each potentially handled by a different actor or system. Operational viability is bounded by the weakest step in the chain, not by the average performance across it. Agentizing a step that is not the bottleneck may improve that step in isolation while leaving overall throughput unchanged. The practical implication is that identifying where the actual constraint lies---whether in user input quality, reviewer capacity, a legacy system integration, or elsewhere---is a prerequisite to any agency or autonomy design decision. Investment directed at the wrong step in the workflow does not raise the potential ceiling regardless of how well the agent performs.

\noindent\textit{\textbf{Evaluation.}}
Unlike the preceding parameters---which determine the ceiling of what is achievable at a given agency--autonomy position---evaluation is the mechanism by which a designer confirms that a chosen configuration performs as intended in a specific deployment. Translating the abstract notion of operational viability into a measurable quantity falls to the system designer. This involves two distinct questions. The first is \emph{what to evaluate}: which dimensions of quality are relevant in the given context---output accuracy, compliance rate, throughput, auditability, operator trust, or error recovery cost---and how to weight them relative to one another. Different stakeholders within the same deployment may prioritise these dimensions differently, and making those priorities explicit is itself a design decision. The second is \emph{how to evaluate}: what measurement infrastructure and methods are required to assess performance against the chosen dimensions---logging, human review sampling, comparison against ground truth, or structured red-teaming for failure modes~\citep{Mazeika2024,Zhu2025}. This is an architectural concern as much as an operational one: a system not designed to produce evaluable outputs cannot be assessed after the fact. Because both questions must be resolved before any output can be measured, none of the design decisions discussed in this work---agency level, autonomy level, or the parameters above---can be validated without first addressing them.

%% file: conclusion.tex
\section{Conclusion}\label{sec:conclusion}

We introduced a two-dimensional design space---autonomy and agency, each organised into five operational levels---as an analytical frame for reasoning about agentic AI deployments in regulated contexts. Drawing on this frame, we proposed six architectural tactics for balancing both axes and grounded the analysis in two worked examples from different deployment domains. We further examined five deployment parameters that shape the ceiling of what is achievable at any given position in the design space independently of the agency--autonomy configuration.

Treating autonomy and agency as explicit, jointly-considered design dimensions makes visible the reliability coupling at their core and the architectural levers available to navigate it. On the autonomy axis, checkpoints and escalation provide mechanisms for reintroducing human oversight at predetermined and conditional points; multi-agent delegation provides the complementary mechanism for raising effective autonomy by replacing a human sub-task handler with a specialised sub-agent. On the agency axis, tool provisioning, tool fencing, and write staging provide mechanisms for adjusting operational reach and reversibility. The six tactics are mechanisms whose selection and combination determine how oversight, reversibility, and accountability are distributed between human and system~\citep{Lange2025}. The design space gives system architects a shared vocabulary for characterising where a deployment sits and a structured set of tactics for reaching a viable configuration within their compliance constraints. The five deployment parameters---model capability, agent architecture, tool fidelity, workflow bottlenecks, and evaluation---situate those design decisions within the broader context of operational constraints that bound what is achievable at any given configuration.

Several limitations bound the current analysis. The iso-viability curves are qualitative reasoning tools whose operationalisation for concrete deployments requires further empirical work, and the set of tactics is illustrative rather than exhaustive. The worked examples are drawn from two deployment contexts; the extent to which the same tactics generalise across different regulatory regimes, organisational types, and task structures remains an open question.

Directions for future work include empirical validation of the design space and tactics across deployment contexts; extension of the analysis to multi-agent workflows in which several specialised agents collaborate within a single service chain; and operationalisation of the iso-viability curves through measurement frameworks that translate qualitative viability into concrete performance thresholds for specific deployment contexts. We believe this work provides a principled foundation for deliberate, compliance-aware design of agentic AI systems---one that future research can build on as deployments in regulated contexts continue to grow in scope and consequence.